\documentclass[runningheads]{llncs}
\usepackage{bbding}
\usepackage{graphicx}
\usepackage{algorithm}
\usepackage{algorithmic}
\usepackage{amsmath,amssymb}
\usepackage{hyperref}
\usepackage[super]{nth}
\usepackage{xcolor}
\usepackage{booktabs}
\usepackage{comment}
\usepackage{float}
\usepackage{subcaption}

\newcommand\Tau{\mathcal{T}}

\begin{document}
\title{Diversity-based Trajectory and Goal Selection with Hindsight Experience Replay}
\titlerunning{Diversity-based Trajectory and Goal Selection with HER}

%\orcidID{0000-1111-2222-3333}

\author{Tianhong Dai\inst{1(}\Envelope\inst{)} \and
Hengyan Liu\inst{1} \and
Kai Arulkumaran\inst{1,2} \and Guangyu Ren\inst{1} \and Anil Anthony Bharath\inst{1}}
\authorrunning{Dai et al.}
\institute{Imperial College London, London SW7 2AZ, UK \\
\email{\{tianhong.dai15, hengyan.liu15, g.ren19, a.bharath\}@imperial.ac.uk} \and
Araya Inc., Tokyo 107-6024, Japan\\
\email{kai\_arulkumaran@araya.org}}
%
%\thanks{This is the preprint presented at PRICAI-2021.}
\maketitle              % typeset the header of the contribution
%
%\vspace{-3mm}
\begin{abstract}
Hindsight experience replay (HER) is a goal relabelling technique typically used with off-policy deep reinforcement learning algorithms to solve goal-oriented tasks; it is well suited to robotic manipulation tasks that deliver only sparse rewards. In HER, both trajectories and transitions are sampled uniformly for training. However, not all of the agent's experiences contribute equally to training, and so naive uniform sampling may lead to inefficient learning. In this paper, we propose diversity-based trajectory and goal selection with HER (DTGSH). Firstly, trajectories are sampled according to the diversity of the goal states as modelled by determinantal point processes (DPPs). Secondly, transitions with diverse goal states are selected from the trajectories by using $k$-DPPs. We evaluate DTGSH on five challenging robotic manipulation tasks in simulated robot environments, where we show that our method can learn more quickly and reach higher performance than other state-of-the-art approaches on all tasks.
%\keywords{Deep Reinforcement Learning  \and Determinantal Point Processes  \and Hindsight Experience Replay.}
\let\thefootnote\relax\footnote{* {This is the preprint presented at PRICAI-2021. The final authenticated publication is available online at \url{https://doi.org/10.1007/978-3-030-89370-5_3}.}}
\end{abstract}
%\vspace{-3mm}
\section{Introduction}
Deep reinforcement learning (DRL)~\cite{arulkumaran2017deep}, in which neural networks are used as function approximators for reinforcement learning (RL), has been shown to be capable of solving complex control problems in several environments, including board games~\cite{schrittwieser2020mastering,silver2017mastering}, video games~\cite{berner2019dota,mnih2015human,vinyals2019grandmaster}, simulated and real robotic manipulation~\cite{andrychowicz2020learning,gu2017deep,levine2016end} and simulated autonomous driving~\cite{kiran2021deep}.

However, learning from a \textit{sparse} reward signal, where the only reward is provided upon the completion of a task, still remains difficult. An agent may rarely or never encounter positive examples from which to learn in a sparse-reward environment. Many domains therefore provide dense reward signals~\cite{brockman2016openai}, or practitioners may turn to reward shaping~\cite{ng1999theory}. Designing dense reward functions typically requires prior domain knowledge, making this approach difficult to generalise across different environments.

Fortunately, a common scenario is goal-oriented RL, where the RL agent is tasked with solving different goals within the same environment~\cite{kaelbling1993learning,schaul2015universal}. Even if each task has a sparse reward, the agent ideally \textit{generalises} across goals, making the learning process easier. For example, in a robotic manipulation task, the goal during a single episode would be to achieve a specific position of a target object.

Hindsight experience replay (HER)~\cite{NIPS2017_453fadbd} was proposed to improve the learning efficiency of goal-oriented RL agents in sparse reward settings: when past experience is replayed to train the agent, the desired goal is replaced (in ``hindsight'') with the achieved goal, generating many positive experiences. In the above example, the desired target position would be overwritten with the achieved target position, with the achieved reward also being overwritten correspondingly.

\begin{figure}[h]
    \centering
    \includegraphics[width=\textwidth]{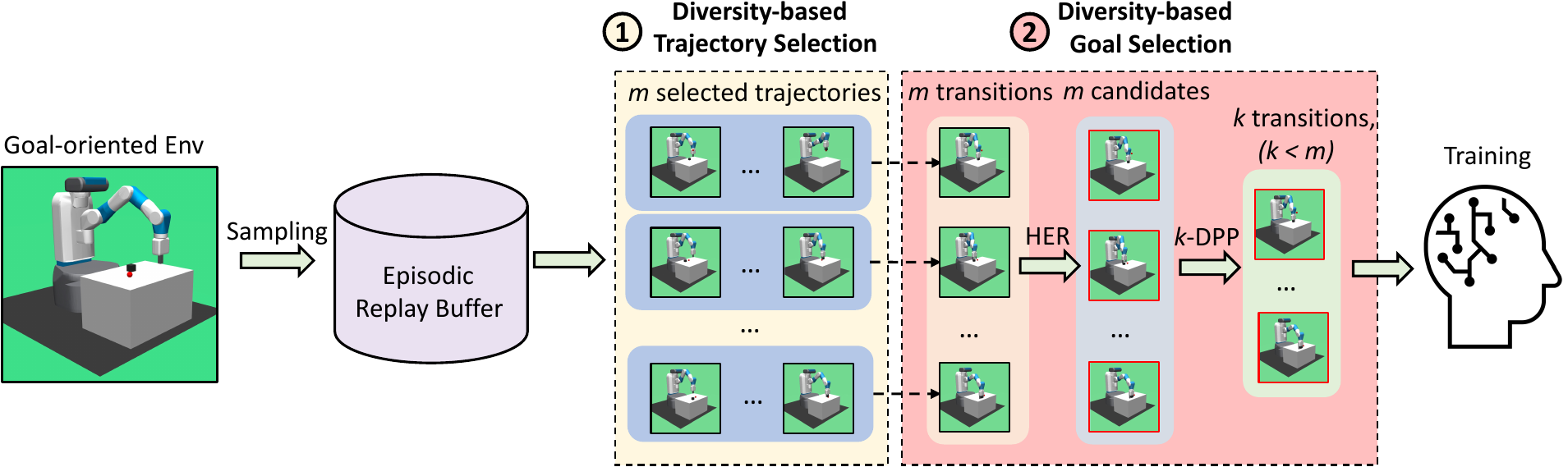}
    \caption{Overview of DTGSH. Every time a new episode is completed, its diversity is calculated, and it is stored in the episodic replay buffer. During training, $m$ episodes are sampled according to their diversity-based priority, and then $k$ diverse, hindsight-relabelled transitions are sampled using a $k$-DPP \cite{kulesza2011k}.}
    \label{fig:illustration}
\end{figure}

We note that HER, whilst it enabled solutions to previously unsolved tasks, can be somewhat inefficient in its use of uniformly sampling transitions during training. In the same way that prioritised experience replay~\cite{schaul2016prioritized} has significantly improved over the standard experience replay in RL, several approaches have improved upon HER by using data-dependent sampling~\cite{fang2019curriculum,zhao2018energy}. HER with energy-based prioritisation (HEBP)~\cite{zhao2018energy} assumes semantic knowledge about the goal-space and uses the energy of the target objects to sample trajectories with high energies, and then samples transitions uniformly. Curriculum-guided HER (CHER)~\cite{fang2019curriculum} samples trajectories uniformly, and then samples transitions based on a mixture of proximity to the desired goal and the diversity of the samples; CHER adapts the weighting of these factors over time. In this work, we introduce diversity-based trajectory and goal selection with HER (DTGSH; See Fig.~\ref{fig:illustration}), which samples trajectories based on the diversity of the goals achieved within the trajectory, and then samples transitions based on the diversity of the set of samples. In this paper, DTGSH is evaluated on five challenging robotic manipulation tasks. From extensive experiments, our proposed method converges faster and reaches higher rewards than prior work, without requiring domain knowledge~\cite{zhao2018energy} or tuning a curriculum~\cite{fang2019curriculum}, and is based on a single concept---determinantal point processes (DPPs)~\cite{kulesza2012determinantal}.

\section{Background}
\subsection{Reinforcement Learning}
RL is the study of agents interacting with their environment in order to maximise their reward, formalised using the framework of Markov decision processes (MDPs)~\cite{sutton2018reinforcement}. At each timestep $t$, an agent receives a state $s_{t}$ from the environment, and then samples an action $a_{t}$ from its policy $\pi(a_{t}|s_{t})$. Next, the action $a_{t}$ is executed in the environment to get the next state $s_{t+1}$, and a reward $r_{t}$. In the episodic RL setting, the objective of the agent is to maximise its expected return $\mathbb{E}[R]$ over a finite trajectory with length $T$:
\begin{equation}
    \mathbb{E}[R] = \mathbb{E}\left[\sum_{t=1}^{T} \gamma^{t-1}r_{t}\right],
\end{equation}
where $\gamma \in [0, 1]$ is a discount factor that exponentially downplays the influence of future rewards, reducing the variance of the return.
%\textbf{Discuss goal-based RL here}

\subsection{Goal-oriented Reinforcement Learning}
RL can be expanded to the multi-goal setting, where the agent's policy and the environment's reward function $\mathcal{R}(s_t, a_t)$ are also conditioned on a goal $g$~\cite{kaelbling1993learning,schaul2015universal}. In this work, we focus on the goal-oriented setting and environments proposed by OpenAI~\cite{plappert2018multi}.

In this setting, every episode comes with a desired goal $g$, which specifies the desired configuration of a target object in the environment (which could include the agent itself). At every timestep $t$, the agent is also provided with the currently achieved goal $g^{ac}_{t+1}$. A transition in the environment is thus denoted as: $(s_t, a_t, r_t, s_{t+1}, g, g^{ac}_{t+1})$. The environment provides a sparse reward function, where a negative reward is given unless the achieved goal is within a small distance $\epsilon$ of the desired goal:
\begin{equation}
\mathcal{R}\left(g, g^{ac}_{t+1}\right):=
\begin{cases}
0& \text{if $\left\|g^{ac}_{t+1} - g\right\|\leq\epsilon$}\\
-1& \text{otherwise}.
\end{cases}
\label{eq:sparse_reward}
\end{equation}

However, in this setting, the agent is unlikely to achieve a non-negative reward through random exploration. To overcome this, HER provides successful experiences for the agent to learn from by relabelling transitions during training: the agent trains on a hindsight desired goal $g'$, which is set to the achieved goal $g^{ac}_{t+1}$, with $r_t$ recomputed using the environment reward function (Eq. \eqref{eq:sparse_reward}).

\subsection{Deep Deterministic Policy Gradient}
Deep deterministic policy gradient (DDPG)~\cite{lillicrapContinuous} is an off-policy actor-critic DRL algorithm for continuous control tasks, and is used as the baseline algorithm for HER \cite{NIPS2017_453fadbd,fang2019curriculum,zhao2018energy}. The actor $\pi_{\theta}(s_{t})$ is a policy network parameterised by $\theta$, and outputs the agent's actions. The critic $Q_{\eta}(s_{t}, a_{t})$ is a state-action-value function approximator parameterised by $\eta$, and estimates the expected return following a given state-action pair. The critic is trained by minimising ${\mathcal{L}_{c}=\mathbb{E}[(Q_{\eta}(s_{t}, a_{t}) - y_{t})^{2}]}$ where ${y_{t} = r_{t} + \gamma Q_{\eta}(s_{t+1}, \pi_{\theta}(s_{t+1}))}$. The actor is trained by maximising ${\mathcal{L}_{a} = \mathbb{E}[Q_{\eta}(s_{t}, \pi_{\theta}(s_{t}))]}$, backpropagating through the critic. Further implementation details can be found in prior work~\cite{NIPS2017_453fadbd,lillicrapContinuous}.

\subsection{Determinantal Point Processes}
A DPP \cite{kulesza2012determinantal} is a stochastic process that characterises a probability distribution over sets of points using the determinant of some function. In machine learning it is often used to quantify the diversity of a subset, with applications such as video~\cite{mahasseni2017unsupervised} and document summarisation~\cite{hong2014improving}.

Formally, for a discrete set of points $\mathcal{Y}=\{x_{1}, x_{2}, \cdots, x_{N}\}$, a point process $\mathcal{P}$ is a probability measure over all $2^{|\mathcal{Y}|}$ subsets. $\mathcal{P}$ is a DPP if a random subset $\mathbf{Y}$ is sampled with probability:
\begin{equation}
    \mathcal{P}_{L}(\mathbf{Y}=Y) = \frac{\text{det}(L_{Y})}{\sum_{Y'\subseteq \mathcal{Y}} \text{det}(L_{Y'})} = \frac{\text{det}(L_{Y})}{\text{det}(L+I)},
\end{equation}
where $Y\subseteq \mathcal{Y}$, $I$ is the identity matrix, $L \in \mathbb{R}^{N\times N}$ is the positive semi-definite DPP kernel matrix, and $L_{Y}$ is the sub-matrix with rows and columns indexed by the elements of the subset $Y$. 

The kernel matrix $L$ can be represented as the Gram matrix $L = X^{T}X$, where each column of $X$ is the feature vector of an item in $\mathcal{Y}$. The determinant, $\text{det}(L_{Y})$, represents the (squared) volume spanned by vectors $x_{i}\in Y$. From a geometric perspective, feature vectors that are closer to being orthogonal to each other will have a larger determinant, and vectors in the spanned subspace are more likely to be sampled: $\mathcal{P}_{L}(\mathbf{Y}=Y) \propto \text{det}(L_{Y})$. Using orthogonality as a measure of diversity, we leverage DPPs to sample diverse trajectories and goals.

\section{Related Work}
The proposed work is built on HER~\cite{NIPS2017_453fadbd} as a way to effectively augment goal-oriented transitions from a replay buffer: to address the problem of sparse rewards, transitions from unsuccessful trajectories are turned into successful ones. HER uses an episodic replay buffer, with uniform sampling over trajectories, and uniform sampling over transitions. However, these samples may be redundant, and many may contribute little to the successful training of the agent.

In the literature, some efforts have been made to increase the efficiency of HER by prioritising more valuable episodes/transitions. Motivated by the work-energy principle in physics, HEBP~\cite{zhao2018energy} assigns higher probability to trajectories in which the target object has higher energy; once the episodes are sampled, the transitions are then sampled uniformly. However, HEBP requires knowing the semantics of the goal space in order to calculate the probability, which is proportional to the sum of the target's potential, kinetic and rotational energies.

CHER~\cite{fang2019curriculum} dynamically controls the sampling of transitions during training based on a mixture of goal proximity and diversity. Firstly, $m$ episodes are uniformly sampled from the episodic replay buffer, and then a minibatch of $k < m$ is sampled according to the current state of the curriculum. The curriculum initially biases sampling to achieved goals that are close to the desired goal (requiring a distance function), and later biases sampling towards diverse goals, using a $k$-nearest neighbour graph and a submodular function to more efficiently sample a diverse subset of goals (using the same distance function).

Other work has expanded HER in orthogonal directions. Hindsight policy gradient \cite{rauber2018hindsight} and episodic self-imitation learning~\cite{dai2020episodic} apply HER to improve the efficiency of goal-based on-policy algorithms. Dynamic HER~\cite{fang2018dher} and competitive ER~\cite{liu2018competitive} expand HER to the dynamic goal and multi-agent settings, respectively.

The use of DPPs in RL has been more limited, with applications towards modelling value functions of sets of agents in multiagent RL~\cite{osogami2019determinantal,pmlr-v119-yang20i}, and most closely related to us, finding diverse policies~\cite{parker2020effective}.

\section{Methodology}
We now formally describe the two main components of our method, DTGSH: 1) a diversity-based trajectory selection module to sample valuable trajectories for the further goal selection; 2) a diversity-based goal selection module to select transitions with diverse goal states from the previously selected trajectories. Together, these select informative transitions from a large area of the goal space, improving the agent's ability to learn and generalise.

\subsection{Diversity-based Trajectory Selection}

We propose a diversity-based prioritization method to select valuable trajectories for efficient training. Related to HEBP's prioritisation of high-energy trajectories~\cite{zhao2018energy}, we hypothesise that trajectories that achieve diverse goal states $g^{ac}_{t}$ are more valuable for training; however, unlike HEBP, we do not require knowledge of the goal space semantics.

In a robotic manipulation task, the agent needs to move a target object from its initial position, $g^{ac}_{1}$, to the target position, $g$. If the agent never moves the object, despite hindsight relabelling it will not be learning information that would directly help in task completion. On the other hand, if the object moves a lot, hindsight relabelling will help the agent learn about meaningful interactions.

In our approach, DPPs are used to model the diversity of achieved goal states $g^{ac}_{t}$ in an episode, or subsets thereof. For a single trajectory $\Tau$ of length $T$, we divide it into several partial trajectories $\tau_{j}$ of length $b$, with achieved goal states $\{g^{ac}_{t}\}_{t=n:n+b-1}$. That is, with a sliding window of $b = 2$, a trajectory $\Tau$ can be divided into $N_p$ partial trajectories:
\begin{equation}
    \Tau_{i} = \{\{\underbrace{g^{ac}_{1}, g^{ac}_{2}}_{\tau_{1}}\}, \{\underbrace{g^{ac}_{2}, g^{ac}_{3}}_{\tau_{2}}\}, \{\underbrace{g^{ac}_{3}, g^{ac}_{4}}_{\tau_{3}}\}, \cdots, \{\underbrace{g^{ac}_{T-1}, g^{ac}_{T}}_{\tau_{N_{p}}}\}\}.
\end{equation}
The diversity $d_{\tau_{j}}$ of each partial trajectory $\tau_{j}$ can be computed as:
\begin{equation}
    d_{\tau_{j}} = \text{det}(L_{\tau_{j}}), 
\label{eq:partial_div}
\end{equation}
where $L_{\tau_{j}}$ is the kernel matrix of partial trajectory $\tau_{j}$:
\begin{equation}
    L_{\tau_{j}} = M^{T}M,
\end{equation}
and $M=[\hat{g}^{ac}_{n}, \hat{g}^{ac}_{n+1}, \cdots, \hat{g}^{ac}_{n+b-1}]$, where each $\hat{g}^{ac}$ is the $\ell_2$-normalised version of the achieved goal $g^{ac}$~\cite{kulesza2011k}. Finally, the diversity $d_\Tau$ of trajectory $\Tau$ is the sum of the diversity of its $N_p$ constituent partial trajectories:
\begin{equation}
    d_\Tau = \sum_{j=1}^{N_{p}} d_{\tau_{j}}.
\label{eq:total_div}
\end{equation}

Similarly to HEBP~\cite{zhao2018energy}, we use a non-uniform episode sampling strategy. During training, we prioritise sampling episodes proportionally to their diversity; the probability $p(\Tau_{i})$ of sampling trajectory $\Tau_{i}$ from a replay buffer of size $N_{e}$ is:
\begin{equation}
    p(\Tau_{i}) = \frac{d_{\Tau_{i}}}{\sum_{n=1}^{N_{e}} d_{\Tau_{n}}}.
\label{eq:diversity_priority}
\end{equation}

\subsection{Diversity-based Goal Selection}
\begin{figure}[h]
    \begin{subfigure}{\textwidth}
        \centering
        \includegraphics[width=0.9\textwidth]{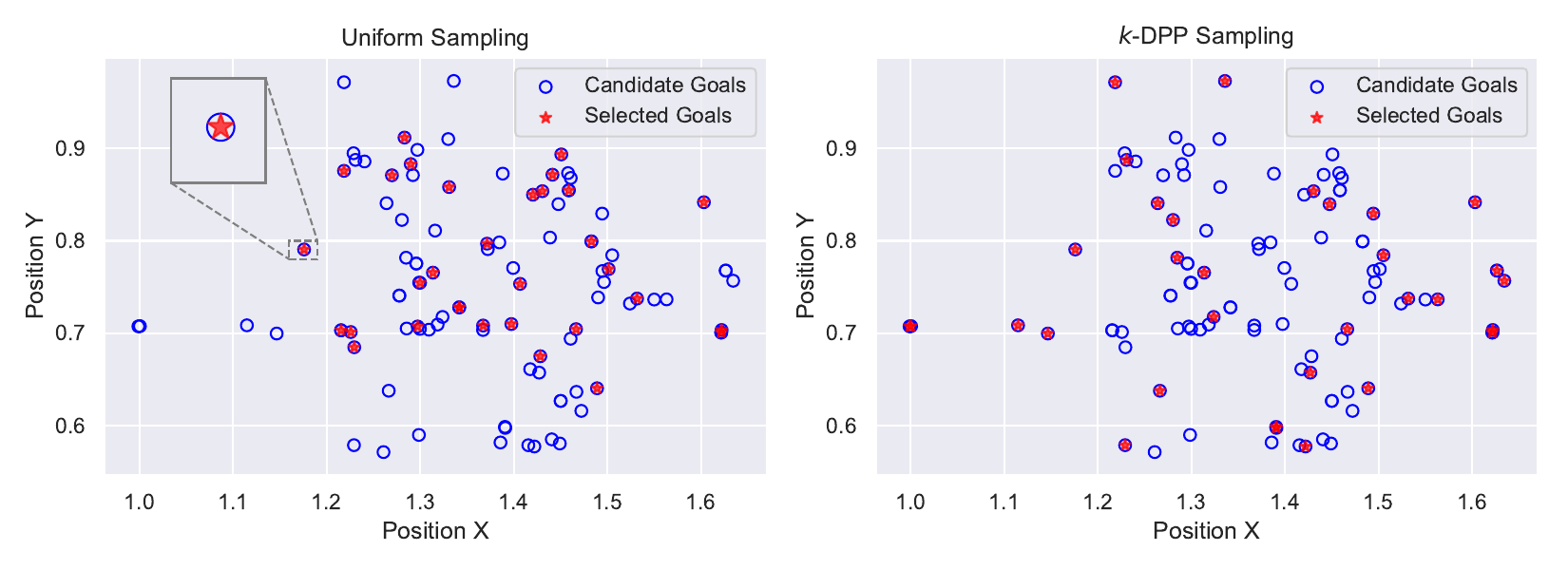}
        \caption{Plot of candidate goals and selected goals. $k$-DPP sampling is more likely to sample points from the full span of the goal space.}
        \label{fig:goal_sampling}
    \end{subfigure}
    \begin{subfigure}{\textwidth}
        \centering
        \includegraphics[width=0.9\textwidth]{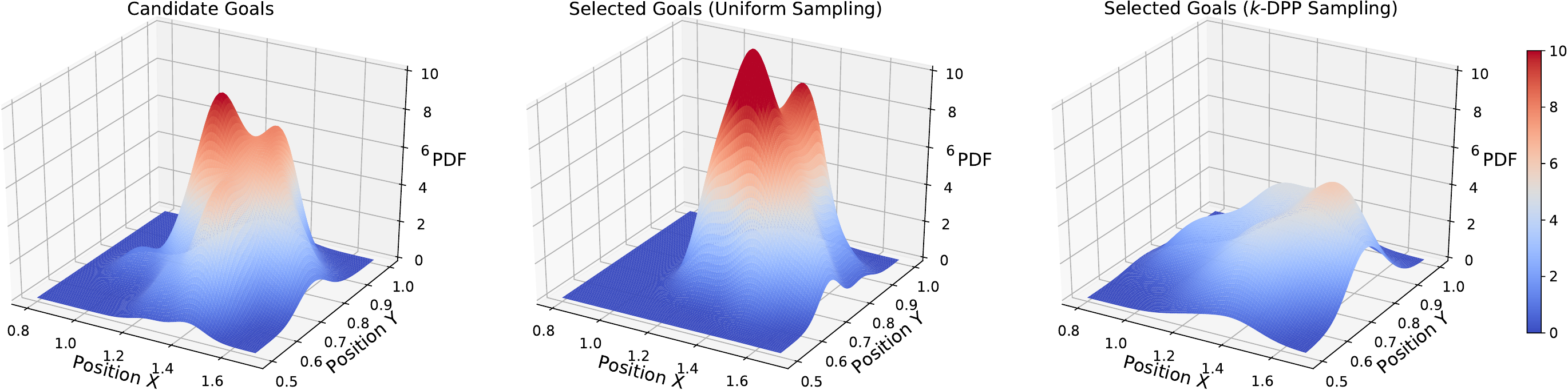}
        \caption{Kernel density estimation of the distributions of goals. $k$-DPP leads to a more uniform selection of goals over the support of the goal space.}
        \label{fig:kde}
    \end{subfigure}
    \caption{Visualisation of $k=32$ goals selected from $m=100$ candidate goals of the Push task using either uniform sampling or $k$-DPP sampling, respectively. The candidate goals are distributed over a 2D ($xy$) space. Note that  $k$-DPP sampling (right hand plots) results in a broader span of selected goals in $xy$ space compared to uniform sampling (left hand plots).}
\end{figure}

In prior work~\cite{NIPS2017_453fadbd,zhao2018energy}, after selecting the trajectories from the replay buffer, one transition from each selected trajectory is sampled uniformly to construct a minibatch for training. However, the modified goals $g^{\prime}$ in the minibatch might be similar, resulting in redundant information. In order to form a minibatch with diverse goals for more efficient learning, we use $k$-DPPs~\cite{kulesza2011k} for sampling goals. Compared to the standard DPP, a $k$-DPP is a conditional DPP where the subset $Y$ has a fixed size $k$, with the probability distribution function:
\begin{equation}
    \mathcal{P}_{L}^{k}(\mathbf{Y}=Y) = \frac{\text{det}(L_{Y})}{\sum_{|Y^{\prime}|=k} \text{det}(L_{Y^{\prime}})}.
\end{equation}
$k$-DPPs are more appropriate for goal selection with a minibatch of fixed size $k$. Given $m > k$ trajectories sampled from the replay buffer, we first uniformly sample a transition from each of the $m$ trajectories. Finally, a $k$-DPP is used to sample a diverse set of transitions based on the relabelled goals $g'$ (which, in this context, we denote as ``candidate goals''). Fig.~\ref{fig:goal_sampling} gives an example of uniform vs. $k$-DPP sampling, demonstrating the increased coverage of the latter. Fig.~\ref{fig:kde} provides corresponding estimated density plots; note that the density of the $k$-DPP samples is actually more uniform over the \textit{support} of the candidate goal distribution.

% pseudo code of dpp-sampling
\begin{algorithm}[H]
\caption{Diversity-based Goal Selection using $k$-DPP}
\label{alg:goal_sample}
\begin{algorithmic}[1]
 \REQUIRE set of $m$ candidate goal states $\mathcal{G}:=\{g_{i}\}_{i=1:m}$, minibatch size $k$
 \STATE $J \leftarrow\varnothing$, $M \leftarrow  [g_{1}, g_{2}, \cdots, g_{m}]$
 \STATE Calculate the DPP kernel matrix $L_M$
 \STATE $\{\boldsymbol{v}_{n}, \lambda_{n}\} \leftarrow \text{EigenDecomposition}(L_M)$
 %\STATE $e_{k}(\lambda_{1}, \cdots, \lambda_{m}): = \sum\limits_{|J'|=k}\prod_{n\in J'}\lambda_{n}$ \hfill{$\triangleright$ elementary symmetric polynomial: $e^{m}_{k}$}
 \STATE $e_k(\lambda_1,\lambda_2,\dots,\lambda_m) := 
  \sum\limits_{J^{\prime} \subseteq \{1,2,\dots,m\}\atop|J^{\prime}|=k} \prod\limits_{n \in J^{\prime}} \lambda_n$ \hfill{$\triangleright$ elementary symmetric polynomial: $e^{m}_{k}$}
 \FOR{$\mathrm{n} = m, m-1, \cdots, 1$}
 \IF{$u\sim \text{Uniform}[0, 1] < \lambda_{n}\frac{e^{n-1}_{k-1}}{e_{k}^{n}}$}
 \STATE $J \leftarrow J \cup \{n\}$, $k \leftarrow k - 1$
 \IF{$k = 0$}
 \STATE \textbf{break}
 \ENDIF
 \ENDIF
 \ENDFOR
 \STATE $V \leftarrow \{\boldsymbol{v}_{n}\}_{n \in J}$, $B \leftarrow \varnothing$
 \WHILE{$|V| > 0$}
 \STATE Select $g_{i}$ from $\mathcal{G}$ with $p(g_{i}):=\frac{1}{|V|}\sum_{\boldsymbol{v}\in V}(\boldsymbol{v}^{T}\boldsymbol{b}_{i})^{2}$\hfill{$\triangleright$ $\boldsymbol{b}_{i}$ is the $i^{\text{th}}$ standard basis}
 \STATE $B \leftarrow B \cup \{g_{i}\}$
 \STATE $V \leftarrow V_{\perp}$ \hfill{$\triangleright$ an orthonormal basis for the subspace of $V$ orthogonal to $\boldsymbol{b}_{i}$}
 \ENDWHILE
 \RETURN minibatch $B$ with size $k$
\end{algorithmic}
\end{algorithm}
\vspace{-5mm}

\begin{algorithm}[H]
\caption{Diversity-based Trajectory and Goal Selection with HER}
\label{alg:main_alg}
\begin{algorithmic}[1]
 \REQUIRE RL environment with episodes of length $T$, number of episodes $N$, off-policy RL algorithm $\mathbb{A}$, episodic replay buffer $\mathcal{B}$, number of algorithm updates $U$, candidate transitions size $m$, minibatch size $k$
 \STATE Initialize the parameters $\theta$ of all models in $\mathbb{A}$
 \STATE $\mathcal{B}\leftarrow\varnothing$
 \FOR{$\mathrm{i} = 1, 2, \cdots, N$}
 \STATE Sample a desired goal $g$ and an initial state $s_{0}$\hfill{$\triangleright$ start a new episode}
 \FOR{$\mathrm{t} = 1, 2, \cdots, T$}
 \STATE Sample an action $a_{t}$ using the policy $\pi(s_{t}, g;\theta)$
 \STATE Execute action $a_{t}$ and get the next state $s_{t+1}$ and achieved goal state $g^{ac}_{t+1}$
 \STATE Calculate $r_{t}$ according to Eq.~\eqref{eq:sparse_reward}
 \STATE Store transition $(s_t, a_t, r_t, s_{t+1}, g, g^{ac}_{t+1})$ in $\mathcal{B}$
 \ENDFOR
 \STATE Calculate the diversity score of current episode $d_{\Tau_{i}}$ using Eq.~\eqref{eq:partial_div} and Eq.~\eqref{eq:total_div}
 \STATE Calculate the diversity-based priority $p(\Tau)$ of each episode in $\mathcal{B}$ using Eq.~\eqref{eq:diversity_priority}
 \FOR{$\mathrm{iteration} = 1, 2, \cdots, U$}
 \STATE Sample $m$ trajectories from $\mathcal{B}$ according to priority $p(\Tau)$
 \STATE Uniformly sample one transition from each of the $m$ trajectories
 \STATE Relabel goals in each transition and recompute the reward to get $m$ candidate transitions $\{(s_{t}, a_{t}, r_{t}^{\prime}, s_{t+1}, g^{\prime})_{n}\}_{n=1:m}$
 \STATE Sample minibatch $B$ with size $k$ from the $m$ candidates using Alg.~\ref{alg:goal_sample}
 \STATE Optimise $\theta$ with minibatch $B$
 \ENDFOR
 \ENDFOR 
\end{algorithmic}
\end{algorithm}

Alg.~\ref{alg:goal_sample} shows the details of the goal selection subroutine, and Alg.~\ref{alg:main_alg} gives the overall algorithm for our method, DTGSH.

\section{Experiments}
\begin{figure}
  \begin{subfigure}[t]{0.19\textwidth}
    \includegraphics[width=\textwidth]{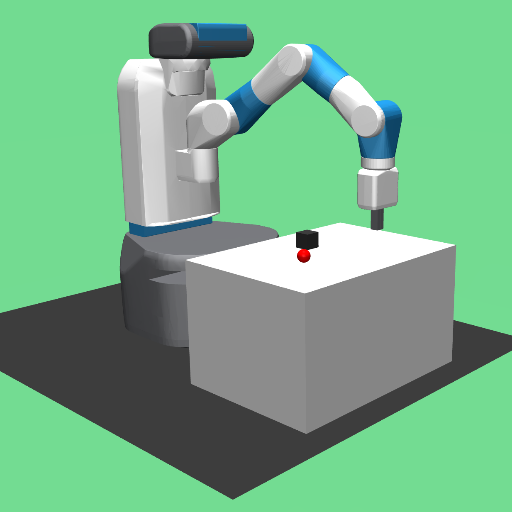}
    \caption{Push}
    \label{subfig:env_push}
  \end{subfigure}\hfill
  \begin{subfigure}[t]{0.19\textwidth}
    \includegraphics[width=\textwidth]{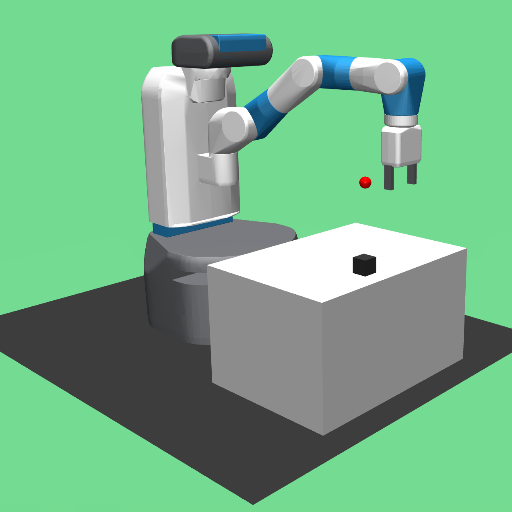}
    \caption{Pick$\&$Place}
    \label{subfig:env_pick}
  \end{subfigure}\hfill
  \begin{subfigure}[t]{0.19\textwidth}
    \includegraphics[width=\textwidth]{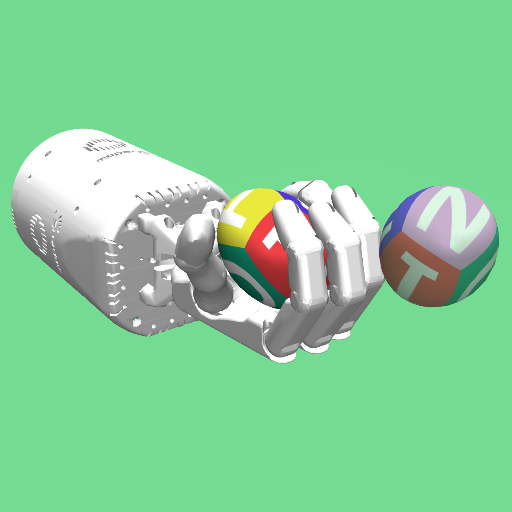}
    \caption{EggFull}
    \label{subfig:env_handegg}
  \end{subfigure}\hfill
  \begin{subfigure}[t]{0.19\textwidth}
    \includegraphics[width=\textwidth]{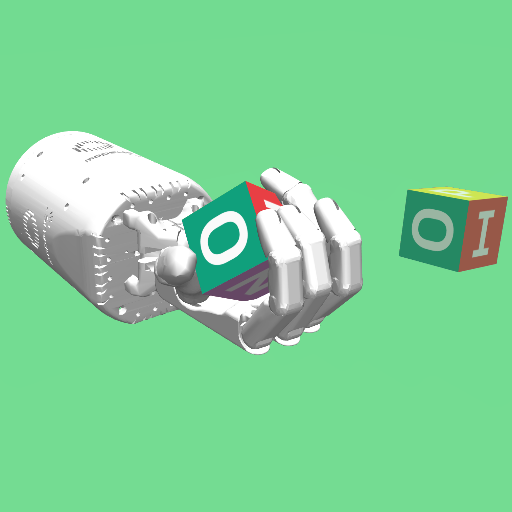}
    \caption{BlockRotate}
    \label{subfig:env_handblock}
  \end{subfigure}\hfill
  \begin{subfigure}[t]{0.19\textwidth}
    \includegraphics[width=\textwidth]{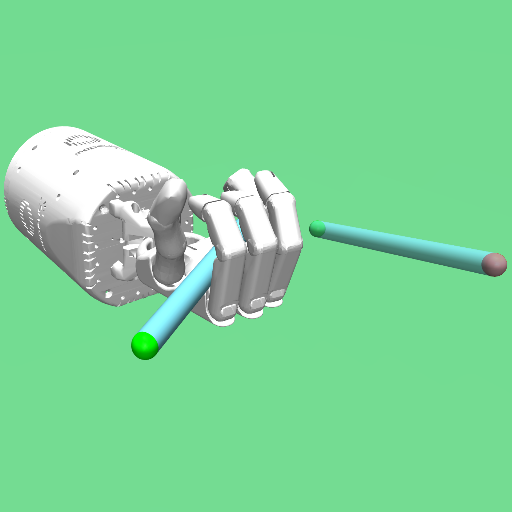}
    \caption{PenRotate}
    \label{subfig:env_handpen}
  \end{subfigure}\hfill
  \caption{Robotic manipulation environments. (a-b) use the Fetch robot, and (c-e) use the Shadow Dexterous Hand.} 
  \label{fig:env}
\end{figure}

We evaluate our proposed method, and compare it with current state-of-the-art HER-based algorithms~\cite{NIPS2017_453fadbd,fang2019curriculum,zhao2018energy} on challenging robotic manipulation tasks~\cite{plappert2018multi}, pictured in Fig.~\ref{fig:env}. Furthermore, we perform ablation studies on our diversity-based trajectory and goal selection modules. Our code is based on OpenAI Baselines\footnote[1]{\url{https://github.com/openai/baselines}}, and is available at: \url{https://github.com/TianhongDai/div-hindsight}.

\subsection{Environments}
The robotic manipulation environments used for training and evaluation include five different tasks. Two tasks use the 7-DoF Fetch robotic arm with two-fingers parallel gripper: Push, and Pick\&Place, which both require the agent to move a cube to the target position. The remaining three tasks use a 24-DoF Shadow Dexterous Hand to manipulate an egg, a block and a pen, respectively. The sparse reward function is given by Eq.~\eqref{eq:sparse_reward}.

In the Fetch environments, the state $s_{t}$ contains the position and velocity of the joints, and the position and rotation of the cube. Each action $a_{t}$ is a 4-dimensional vector, with three dimensions specifying the relative position of the gripper, and the final dimension specifying the state of the gripper (i.e., open or closed). The desired goal $g$ is the target position, and the achieved goal $g^{ac}_{t}$ is the position of the cube. Each episode is of length $T = 50$.

In the Shadow Dexterous Hand environments, the state $s_{t}$ contains the position and velocity of the joints. Each action $a_{t}$ is a 20-dimensional vector which specifies the absolute position of 20 non-coupled joints in the hand. The desired goal $g$ and achieved goal $g^{ac}_t$ specify the rotation of the object for the block and pen tasks, and the position + rotation of the object for the egg task. Each episode is of length $T = 200$.

\subsection{Training Settings}
We base our training setup on CHER~\cite{fang2019curriculum}. We train all agents on minibatches of size $k = 64$ for 50 epochs using MPI for parallelisation over 16 CPU cores; each epoch consists of 1600 ($16 \times 100$) episodes, with evaluation over 160 ($16 \times 10$) episodes at the end of each epoch. Remaining hyperparameters for the baselines are taken from the original work~\cite{NIPS2017_453fadbd,fang2019curriculum,zhao2018energy}. Our method, DTGSH, uses partial trajectories of length $b = 2$ and $m = 100$ as the number of candidate goals.

\subsection{Benchmark Results}
We compare DTGSH to DDPG~\cite{lillicrapContinuous}, DDPG+HER~\cite{NIPS2017_453fadbd}, DDPG+HEBP~\cite{zhao2018energy} and DDPG+CHER~\cite{fang2019curriculum}. Evaluation results are given based on repeated runs with 5 different seeds; we plot the median success rate with upper and lower bounds given by the \nth{75} and \nth{25} percentiles, respectively.

\begin{figure}
  \begin{subfigure}[t]{0.33\textwidth}
    \includegraphics[width=\textwidth]{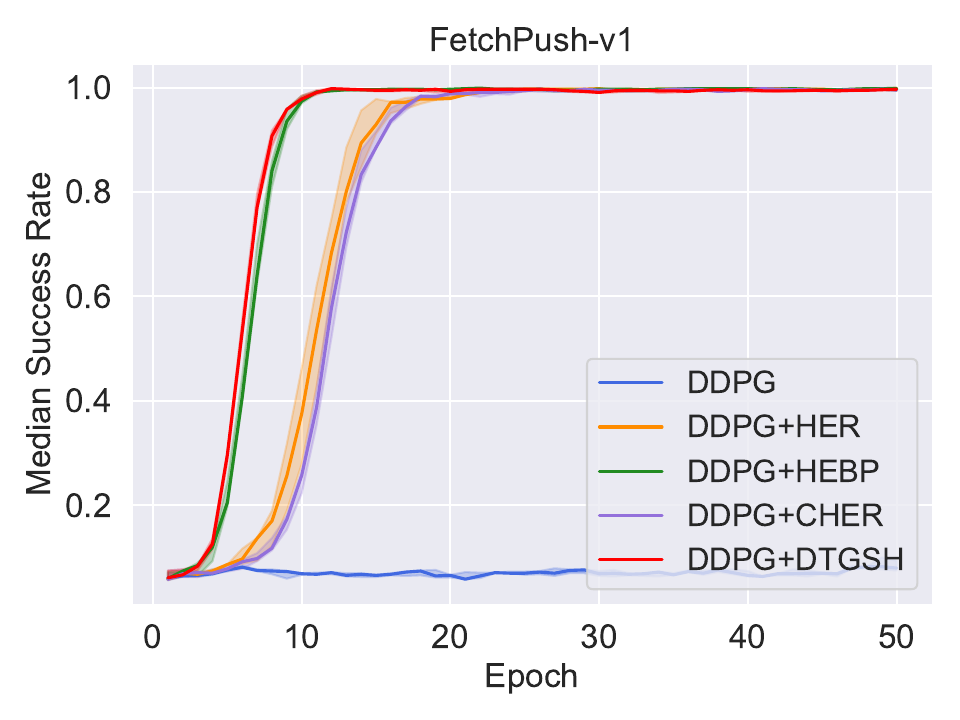}
    \caption{Push}
    \label{subfig:baseline_push}
  \end{subfigure}\hfill
  \begin{subfigure}[t]{0.33\textwidth}
    \includegraphics[width=\textwidth]{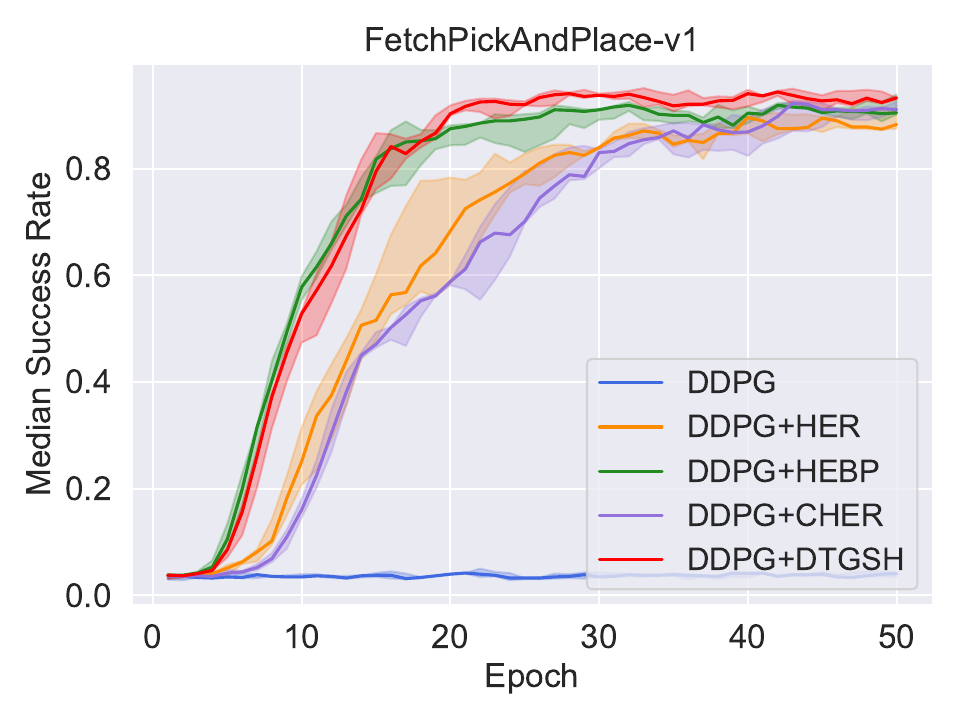}
    \caption{Pick\&Place}
    \label{subfig:baseline_pick}
  \end{subfigure}\hfill
  \begin{subfigure}[t]{0.33\textwidth}
    \includegraphics[width=\textwidth]{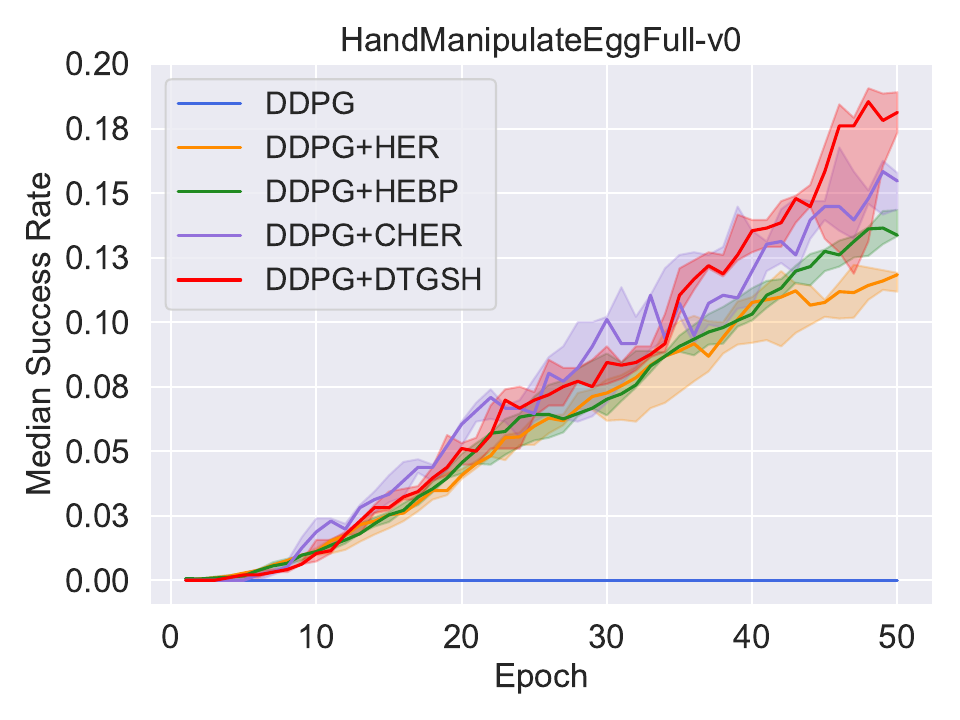}
    \caption{EggFull}
    \label{subfig:baseline_handegg}
  \end{subfigure}\hfill
  \begin{subfigure}[t]{0.33\textwidth}
    \includegraphics[width=\textwidth]{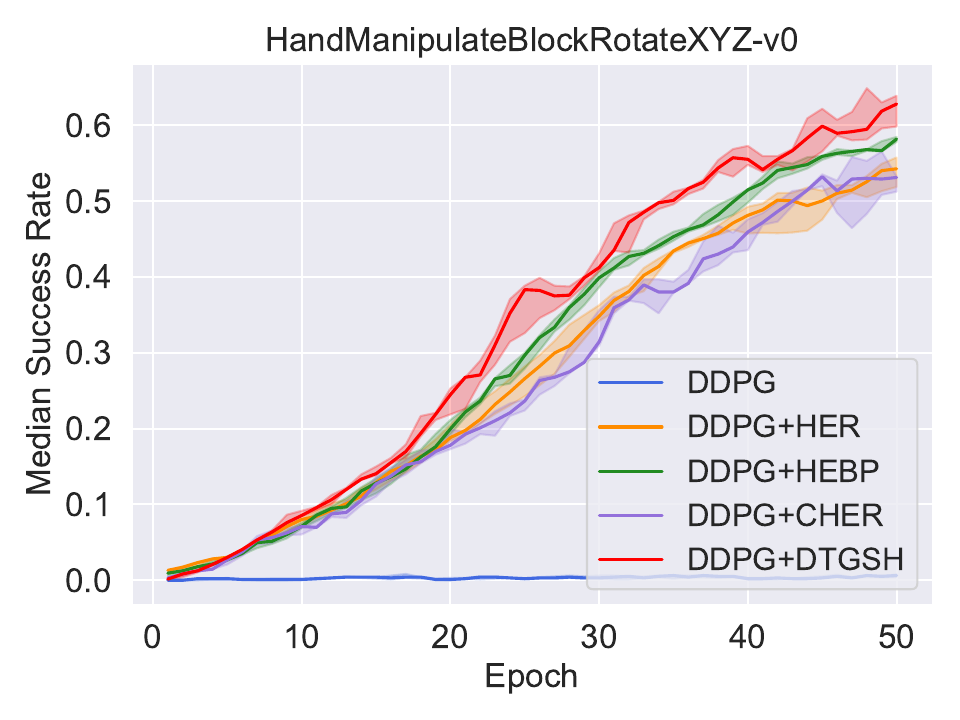}
    \caption{BlockRotate}
    \label{subfig:baseline_handblock}
  \end{subfigure}\hfill
  \begin{subfigure}[t]{0.33\textwidth}
    \includegraphics[width=\textwidth]{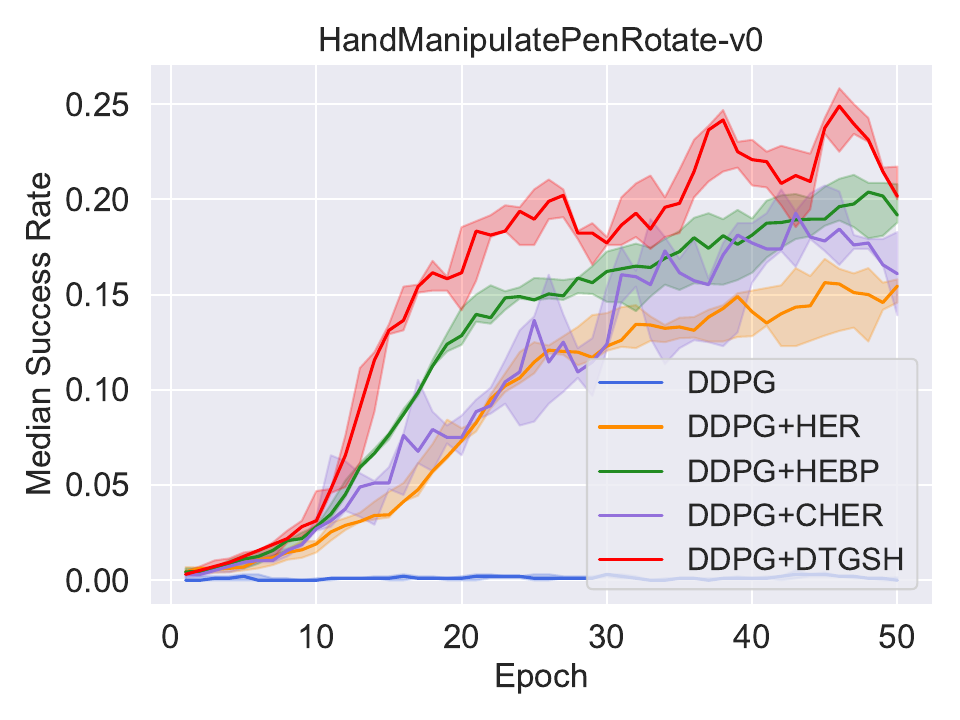}
    \caption{PenRotate}
    \label{subfig:baseline_handpen}
  \end{subfigure}\hfill
  \begin{subfigure}[t]{0.33\textwidth}
    \includegraphics[width=\textwidth]{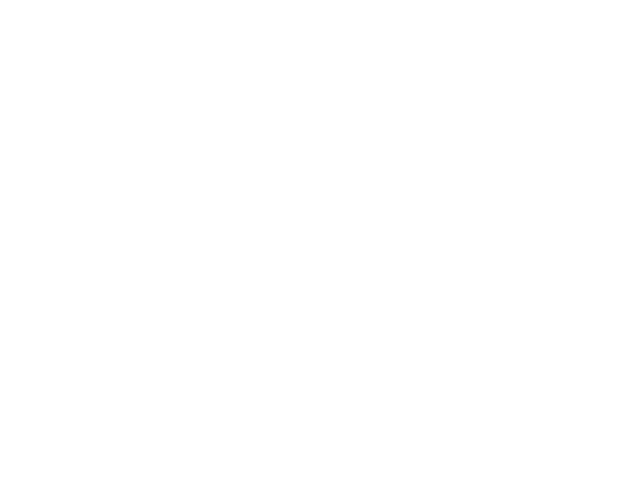}
  \end{subfigure}\hfill
  \caption{Success rate of DTGSH and baseline approaches.} 
  \vspace{-5mm}
  \label{fig:baseline_compare}
\end{figure}

Fig.~\ref{fig:baseline_compare} and Tab.~\ref{tab:benchmark} show the performance of DDPG+DTGSH and baseline approaches on all five tasks. In the Fetch tasks, DDPG+DTGSH and DDPG+HEBP both learn significantly faster than the other methods, while in the Shadow Dexterous Hand tasks DDPG+DTGSH learns the fastest and achieves higher success rates than all other methods. In particular, DDPG cannot solve any tasks without using HER, and CHER performs worse in the Fetch tasks. We believe the results highlight the importance of sampling both diverse trajectories \textit{and} goals, as in our proposed method, DTGSH.
\vspace{-5mm}
\begin{table}[h]
    \centering
    \resizebox{\textwidth}{!}{
    \begin{tabular}{c|c@{\hskip 0.1in}|c@{\hskip 0.1in}|c|c|c}
    \toprule
         & Push & Pick\&Place & EggFull & BlockRotate & PenRotate \\
        \midrule
        DDPG~\cite{lillicrapContinuous} & 0.09$\pm$0.01 & 0.04$\pm$0.00 & 0.00$\pm$0.00 & 0.01$\pm$0.00 & 0.00$\pm$0.00 \\
        DDPG+HER~\cite{NIPS2017_453fadbd} & \textbf{1.00$\pm$0.00} & 0.89$\pm$0.03 & 0.11$\pm$0.01 & 0.55$\pm$0.04 & 0.15$\pm$0.02 \\
        DDPG+HEBP~\cite{zhao2018energy} & \textbf{1.00$\pm$0.00} & 0.91$\pm$0.03 & 0.14$\pm$0.02 & 0.59$\pm$0.02 & 0.20$\pm$0.03\\
        DDPG+CHER~\cite{fang2019curriculum} & \textbf{1.00$\pm$0.00} & 0.91$\pm$0.04 & 0.15$\pm$0.01 & 0.54$\pm$0.04 & 0.17$\pm$0.03 \\
        \midrule
        DDPG+DTGSH & \textbf{1.00$\pm$0.00} & \textbf{0.94$\pm$0.01} & \textbf{0.17$\pm$0.03} & \textbf{0.62$\pm$0.02} & \textbf{0.21$\pm$0.02}\\
        \bottomrule
    \end{tabular}
    }
    \vspace{0.2em}
    \caption{Final mean success rate $\pm$ standard deviation, with best results in \textbf{bold}.}
    \vspace{-14mm}
    \label{tab:benchmark}
\end{table}
%\vspace{-15mm}
\subsection{Ablation Studies}
In this section, we perform the following experiments to investigate the effectiveness of each component in DTGSH: 1) diversity-based trajectory selection with HER (DTSH) and diversity-based goal selection with HER (DGSH) are evaluated independently to assess the contribution of each stage; 2) the performance using different partial trajectory lengths $b$; 3) the performance of using different candidate goal set sizes $m$.
\begin{figure}[h]
  \begin{subfigure}[t]{0.33\textwidth}
    \includegraphics[width=\textwidth]{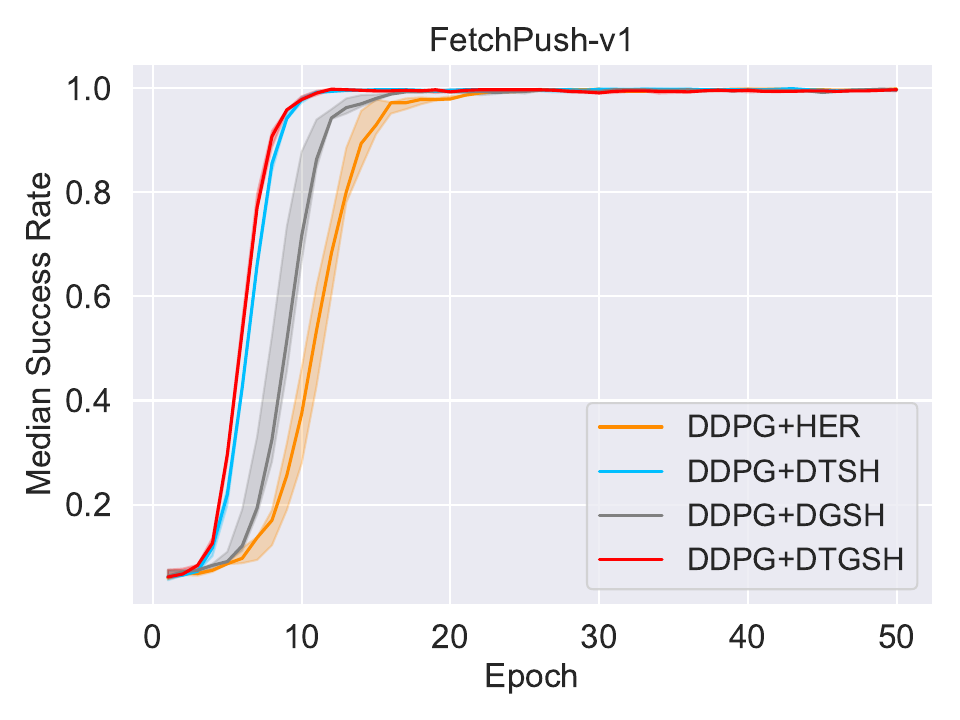}
    \caption{Push}
    \label{subfig:baseline_push_ab1}
  \end{subfigure}\hfill
  \begin{subfigure}[t]{0.33\textwidth}
    \includegraphics[width=\textwidth]{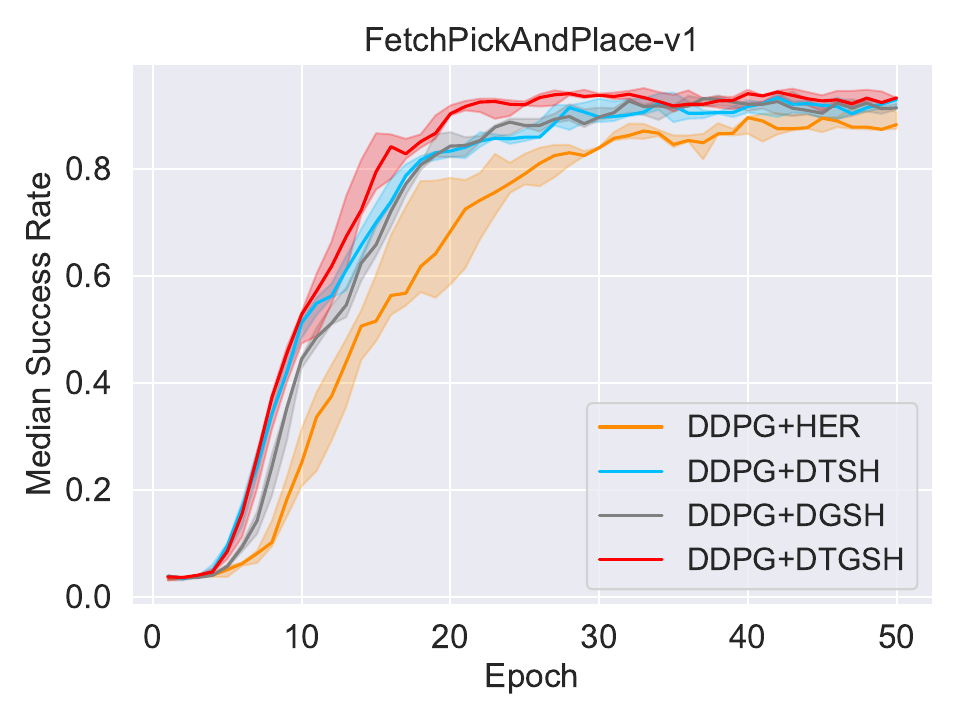}
    \caption{Pick\&Place}
    \label{subfig:baseline_pick_ab1}
  \end{subfigure}\hfill
  \begin{subfigure}[t]{0.33\textwidth}
    \includegraphics[width=\textwidth]{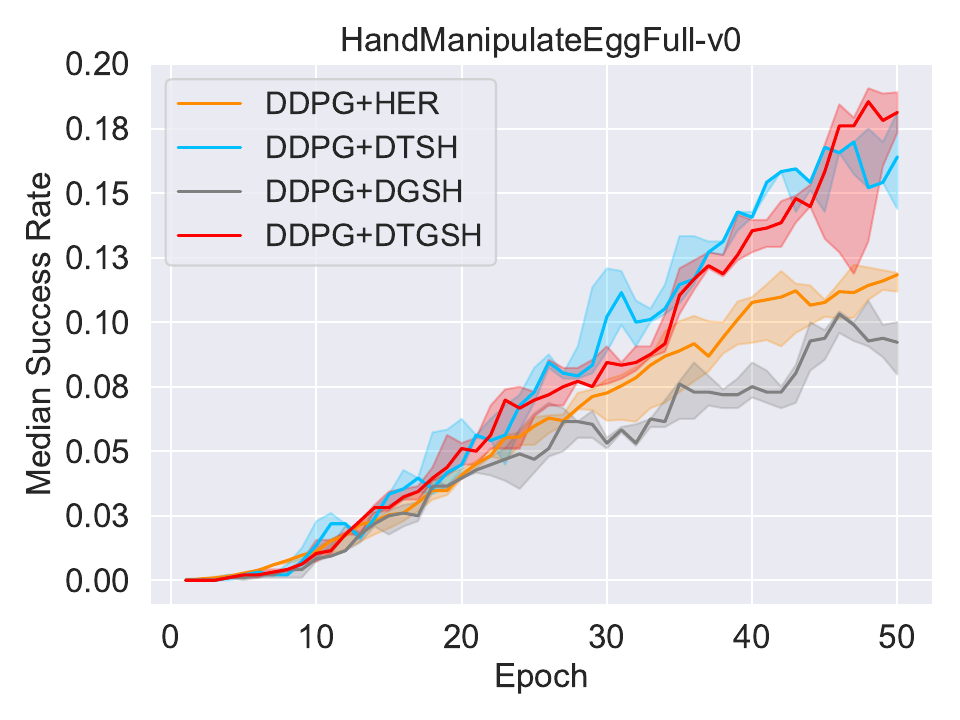}
    \caption{EggFull}
    \label{subfig:baseline_handegg_ab1}
  \end{subfigure}\hfill
  \begin{subfigure}[t]{0.33\textwidth}
    \includegraphics[width=\textwidth]{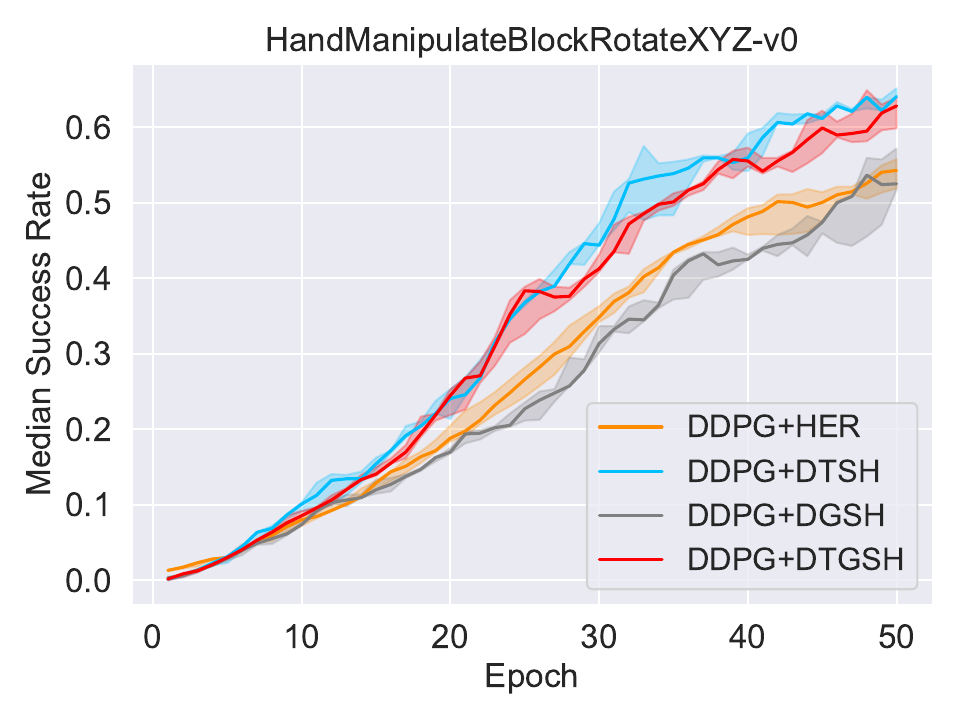}
    \caption{BlockRotate}
    \label{subfig:baseline_handblock_ab1}
  \end{subfigure}\hfill
  \begin{subfigure}[t]{0.33\textwidth}
    \includegraphics[width=\textwidth]{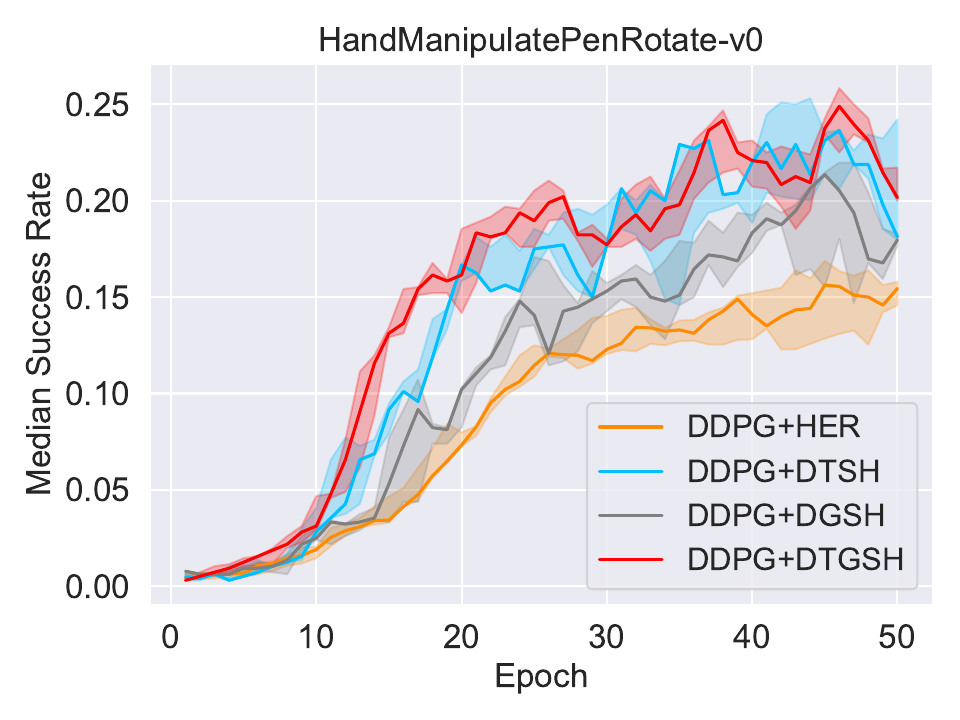}
    \caption{PenRotate}
    \label{subfig:baseline_handpen_ab1}
  \end{subfigure}\hfill
  \begin{subfigure}[t]{0.33\textwidth}
    \includegraphics[width=\textwidth]{imgs/blank.png}
  \end{subfigure}\hfill
  \caption{Success rate of HER, DTGSH, and ablations DTSH and DGSH.} 
  \label{fig:ablation1}
  \vspace{-7mm}
\end{figure}

Fig.~\ref{fig:ablation1} shows the performance of using DTSH and DGSH independently. DDPG+DTSH outperforms DDPG+HER substantially in all tasks, which supports the view that sampling trajectories with diverse achieved goals can substantially improve performance. Furthermore, unlike DDPG+HEBP, DTSH does not require knowing the structure of the goal space in order to calculate the energy of the target object; DDPG+DGSH achieves better performance than DDPG+HER in three environments, and is only worse in one environment. DGSH performs better in environments where it is easier to solve the task (e.g., Fetch tasks), and hence the trajectories selected are more likely to contain useful transitions. However, DTGSH, which is the combination of both modules, performs the best overall.
%\vspace{-1mm}

\begin{figure}[H]
  \begin{subfigure}[t]{0.33\textwidth}
    \includegraphics[width=\textwidth]{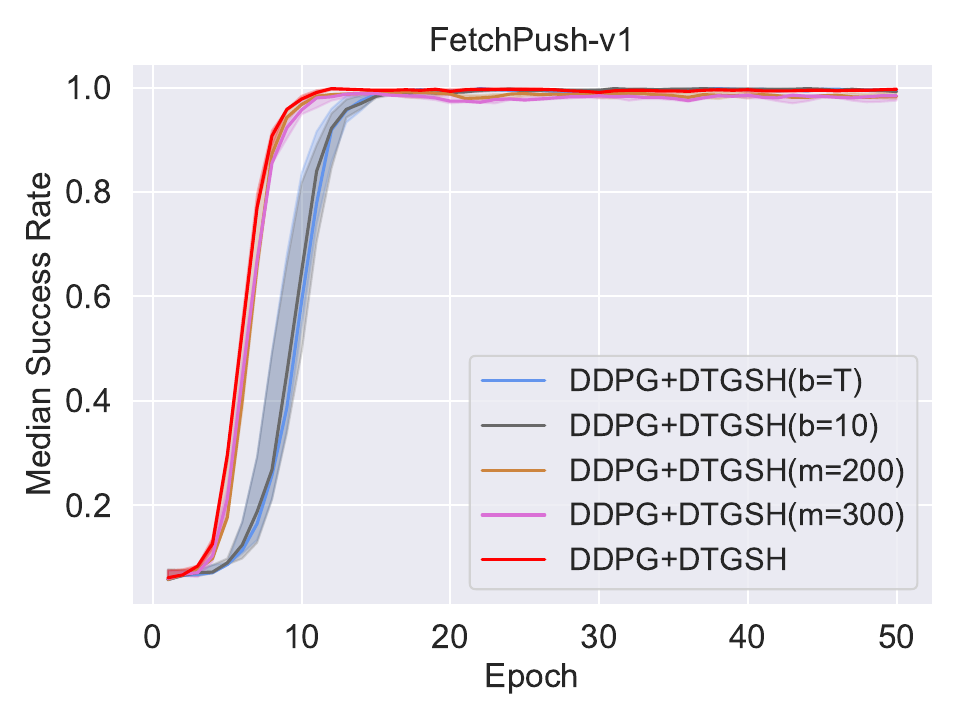}
    \caption{Push}
    \label{subfig:baseline_push_ab2}
  \end{subfigure}\hfill
  \begin{subfigure}[t]{0.33\textwidth}
    \includegraphics[width=\textwidth]{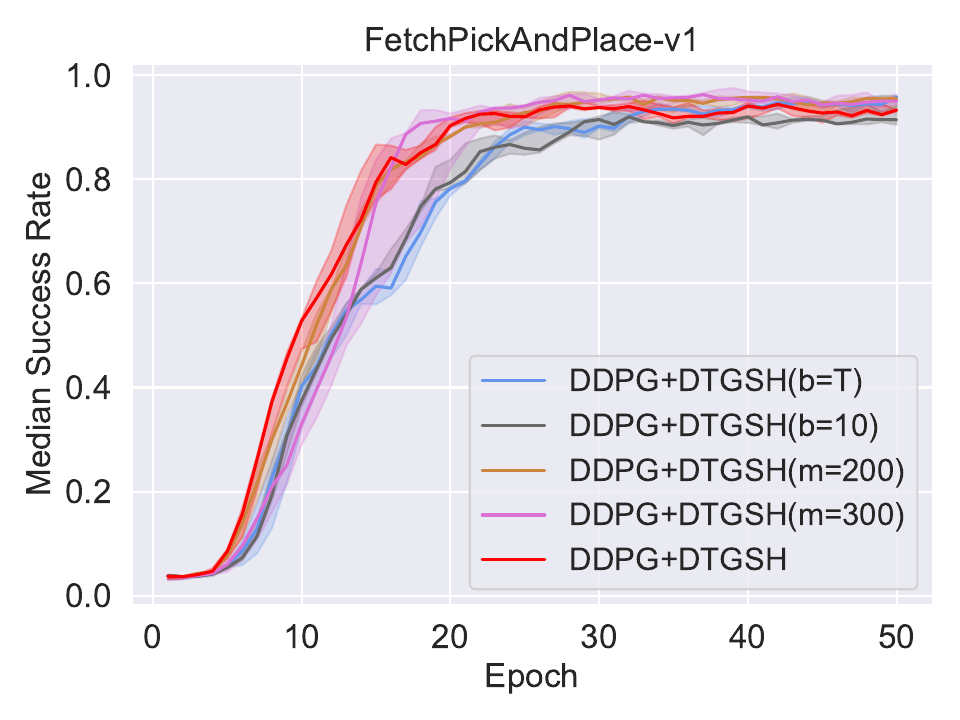}
    \caption{Pick\&Place}
    \label{subfig:baseline_pick_ab2}
  \end{subfigure}\hfill
  \begin{subfigure}[t]{0.33\textwidth}
    \includegraphics[width=\textwidth]{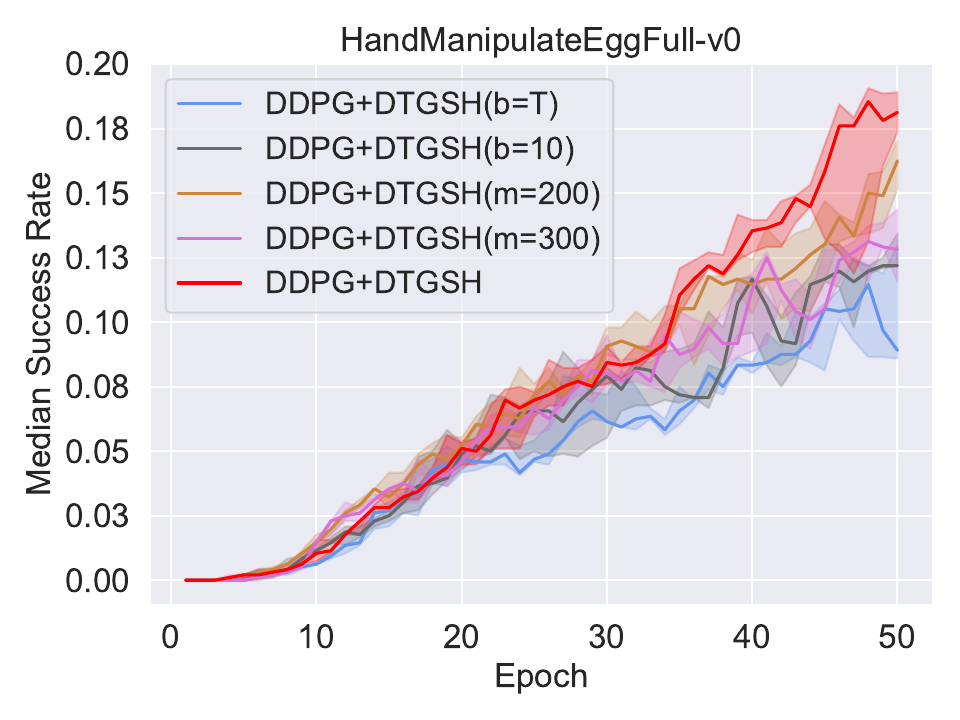}
    \caption{EggFull}
    \label{subfig:baseline_handegg_ab2}
  \end{subfigure}\hfill
  \begin{subfigure}[t]{0.33\textwidth}
    \includegraphics[width=\textwidth]{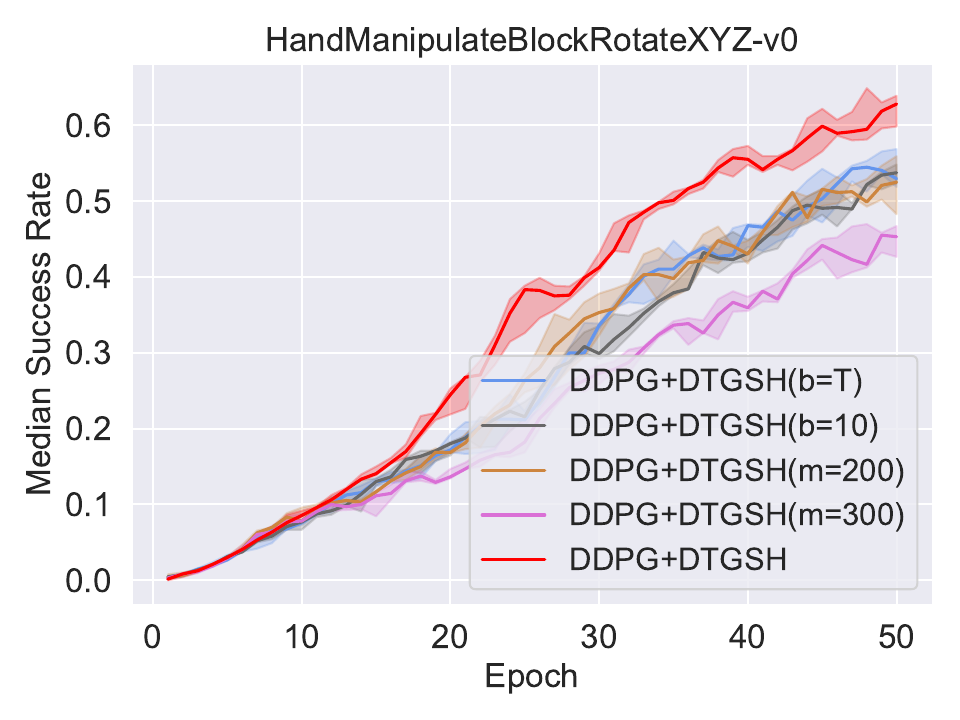}
    \caption{BlockRotate}
    \label{subfig:baseline_handblock_ab2}
  \end{subfigure}\hfill
  \begin{subfigure}[t]{0.33\textwidth}
    \includegraphics[width=\textwidth]{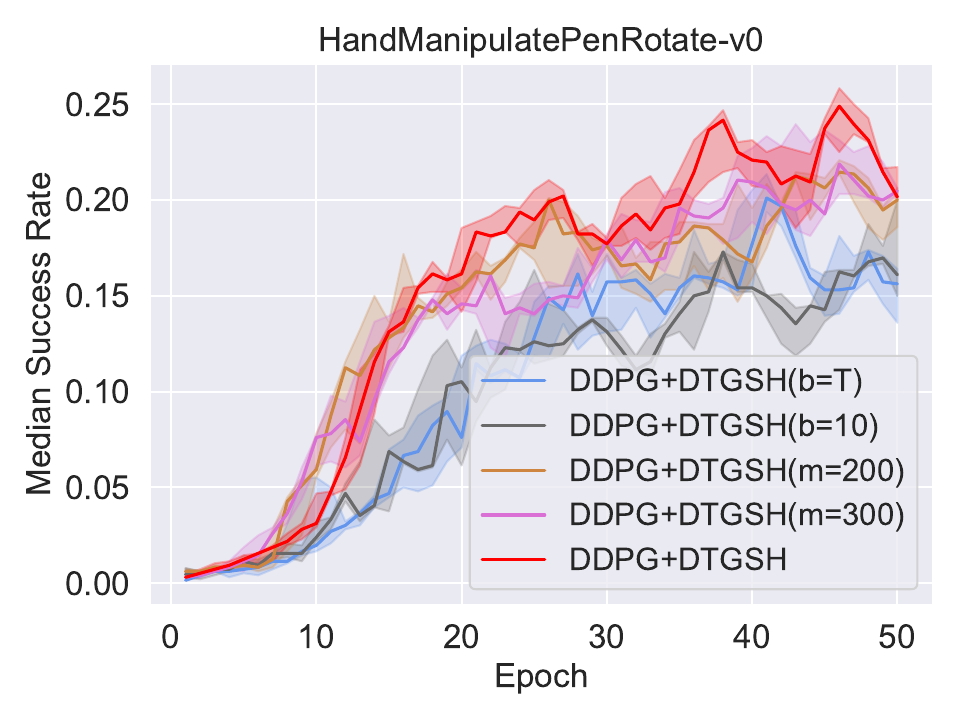}
    \caption{PenRotate}
    \label{subfig:baseline_handpen_ab2}
  \end{subfigure}\hfill
  \begin{subfigure}[t]{0.33\textwidth}
    \includegraphics[width=\textwidth]{imgs/blank.png}
  \end{subfigure}\hfill
  \caption{Success rate of DTGSH with different partial trajectory lengths $b$ and different candidate goal set sizes $m$.} 
  \label{fig:ablation2}
  \vspace{-8mm}
\end{figure}

Fig.~\ref{fig:ablation2} shows the performance of DDPG+DTGSH with different partial trajectory lengths $b$ and different candidate goal set sizes $m$. In this work, we use $b = 2$ and $m = 100$ as the defaults. Performance degrades with $b \gg 2$, indicating that pairwise diversity is best for learning in our method. $m \gg 100$ does not affect performance in the Fetch environments, but degrades performance in the Shadow Dexterous Hand environments.

\subsection{Time Complexity}
Table~\ref{tab:time} gives example training times of all of the HER-based algorithms. DTGSH requires an additional calculation of the diversity score of $\mathcal{O}(N_{p}b^3)$ at the end of every training episode, and sampling of $\mathcal{O}(mk^2)$ for each minibatch.
\begin{table}[h]
    \centering
    \resizebox{\textwidth}{!}{
    \begin{tabular}{c|c|c|c|c}
    \toprule
           & DDPG+HER~\cite{NIPS2017_453fadbd} & DDPG+HEBP~\cite{zhao2018energy} & DDPG+CHER~\cite{fang2019curriculum} & DDPG+DTGSH\\
        \midrule
        Time  & 00:55:08 & 00:56:32 & 03:02:18 & 01:52:30 \\
        \bottomrule
    \end{tabular}
    }
    \vspace{0.2em}
    \caption{Training time (hours:minutes:seconds) of DTGSH and baseline approaches on the Push task for 50 epochs.}
    \vspace{-14mm}
    \label{tab:time}
\end{table}

\section{Conclusion}
In this paper, we introduced diversity-based trajectory and goal selection with hindsight experience replay (DTGSH) to improve the learning efficiency of goal-orientated RL agents in the sparse reward setting. Our method can be divided into two stages: 1) valuable trajectories are selected according to diversity-based priority, as modelled by determinantal point processes (DPPs)~\cite{kulesza2012determinantal}; 2) $k$-DPPs~\cite{kulesza2011k} are leveraged to sample transitions with diverse goal states from previously selected trajectories for training. Our experiments empirically show that DTGSH achieves faster learning and higher final performance in five challenging robotic manipulation tasks, compared to previous state-of-the-art approaches~\cite{NIPS2017_453fadbd,fang2019curriculum,zhao2018energy}. Furthermore, unlike prior extensions of hindsight experience replay, DTGSH does not require semantic knowledge of the goal space~\cite{zhao2018energy}, and does not require tuning a curriculum~\cite{fang2019curriculum}.

% comment out before camera-ready
\section*{Acknowledgements}
This work was supported by JST, Moonshot R\&D Grant Number JPMJMS2012.

%\renewcommand{\bibsection}{\section*{References}}
% The instruction said I can only use author-year systems!
\bibliographystyle{splncs04}
\bibliography{arxiv}

\end{document}